\documentclass[letterpaper, 10 pt, conference]{ieeeconf}  

\IEEEoverridecommandlockouts                              
\usepackage{amsmath} 
\usepackage{tabularx}
\usepackage{graphicx}
\usepackage{subcaption}
\usepackage{multirow}
\usepackage{xcolor}
\title{\LARGE \bf
OxIOD: The Dataset for Deep Inertial Odometry
}

\begin{document}

\author{Changhao Chen$^{*}$, Peijun Zhao$^{*}$, Chris Xiaoxuan Lu, Wei Wang, Andrew Markham, Niki Trigoni
\thanks{*The first two authors contributed equally to this work}
\thanks{The authors are with Department of Computer Science,
        University of Oxford, Oxford OX1 3QD, United Kingdom
        ({\tt\small firstname.lastname@cs.ox.ac.uk})}%
}

\maketitle
\thispagestyle{empty}
\pagestyle{empty}

\begin{abstract}

Advances in micro-electro-mechanical (MEMS) techniques enable inertial measurements units (IMUs) to be small, cheap, energy efficient, and widely used in smartphones, robots, and drones. Exploiting inertial data for accurate and reliable navigation and localization has attracted significant research and industrial interest, as IMU measurements are completely ego-centric and generally environment agnostic. Recent studies have shown that the notorious issue of drift can be significantly alleviated by using deep neural networks (DNNs)~\cite{Chen2018}. However, the lack of sufficient labelled data for training and testing various architectures limits the proliferation of adopting DNNs in IMU-based tasks.
In this paper, we propose and release the \textit{Oxford Inertial Odometry Dataset (OxIOD)}, a first-of-its-kind data collection for inertial-odometry research, with all sequences having ground-truth labels. Our dataset contains 158 sequences totalling more than 42 km in total distance, much larger than previous inertial datasets. Another notable feature of this dataset lies in its diversity, which can reflect the complex motions of phone-based IMUs in various everyday usage. The measurements were collected with four different attachments (handheld, in the pocket, in the handbag and on the trolley), four motion modes (halting, walking slowly, walking normally, and running), five different users, four types of off-the-shelf consumer phones, and large-scale localization from office buildings. Deep inertial tracking experiments were conducted to show the effectiveness of our dataset in training deep neural network models and evaluate learning-based and model-based algorithms. The OxIOD Dataset is available at: http://deepio.cs.ox.ac.uk

\end{abstract}

\section{INTRODUCTION}

Performing inertial navigation and localization is a promising research direction with a vast number of applications ranging from robot navigation \cite{Leutenegger2015}, pedestrian navigation \cite{Harle2013}, activity/health monitoring \cite{Gowda2017} to augmented reality \cite{Marchand2016}. Modern micro-electro-mechanical (MEMS) inertial measurements units (IMUs) are small (a few mm$^2$), cheap (several dollars a piece), energy efficient and widely employed in smartphones, robots and drones. Unlike GPS, vision, radio or other sensor modalities, an inertial tracking solution is completely self-contained and suffers less from environmental impact. Hence, exploiting inertial measurements for accurate navigation and localization is of key importance for human and mobile agents. 
In this work we concentrate on inertial odometry for pedestrian tracking, as a promising technique for ubiquitous indoor/outdoor positioning.

	\begin{figure}
    	\centering
        \includegraphics[width=0.3\textwidth]{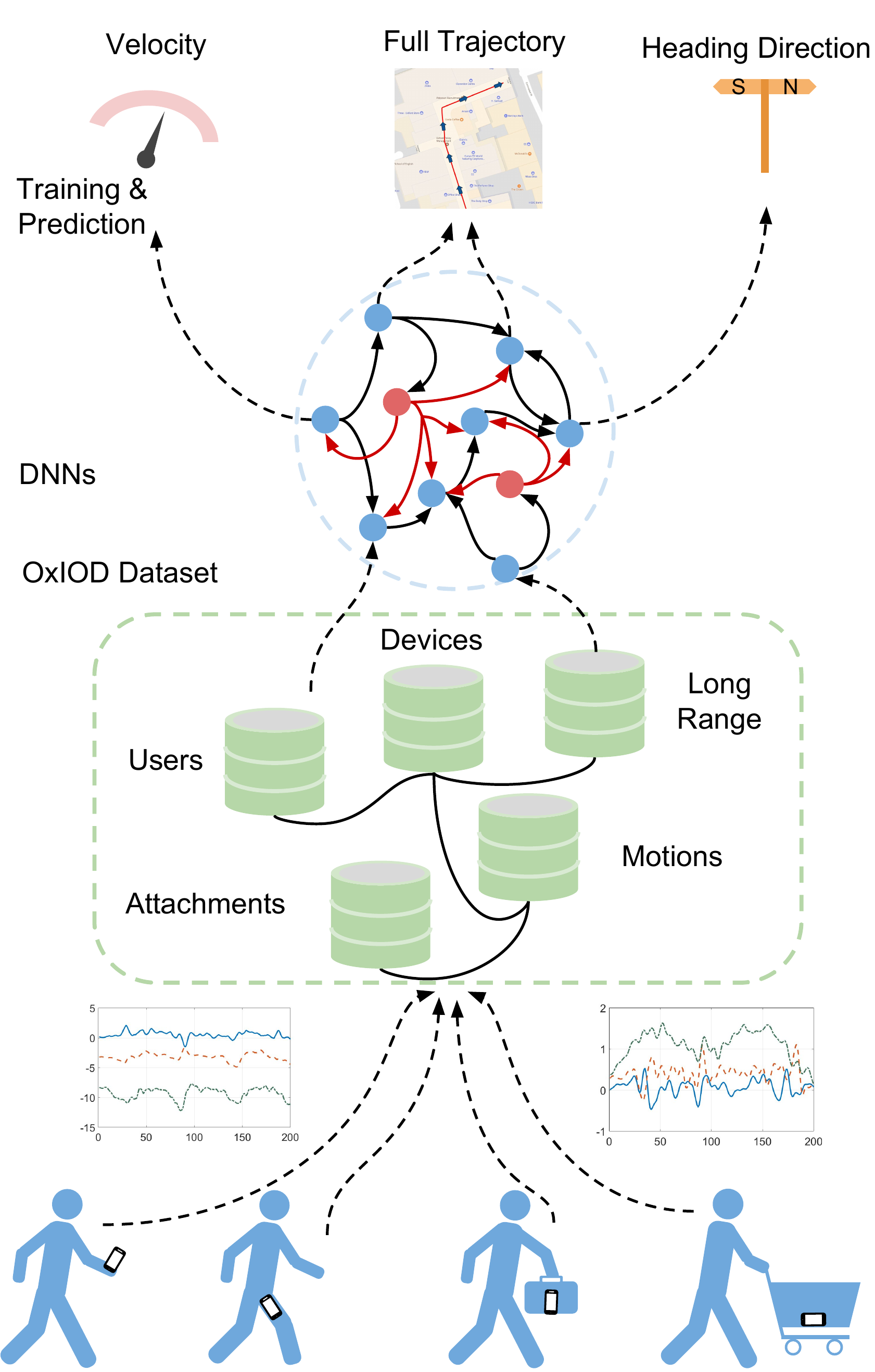}
        \caption{\label{fig:comparison} OxIOD dataset consists of the inertial and magnetic data from mobile devices, suitable for training and evaluating learning-based inertial odometry methods. }
    \end{figure} 

A major drawback and limitation of inertial measurements from low-cost IMUs is the impact of various error sources and biases, leading to unbounded error growth \cite{Naser2008}. One common solution is to combine inertial sensor with camera as visual-inertial-odometry (VIO) \cite{Li2013b} \cite{Leutenegger2015} \cite{Forster2017}, performing well in tracking mobile robots. For pedestrian tracking, IMU has to be attached on users' foot to take advantage of zero-velocity detection and update (ZUPT) for compensating inertial systems drifts \cite{Prieto2016} \cite{Nilsson2012}. Pedestrian dead reckonings (PDRs) \cite{Harle2013} \cite{Xiao2014} were proposed to estimate trajectories by detecting and updating steps. However, relying on strict assumptions, these model-based methods are too restrictive. VIOs require careful time synchronization and calibration, and have to operate in good light conditions. Given personal walking models, ZUPTs and PDRs are assumed to work under periodic pedestrian motion. Hence, the limitations of these handcrafted models prevent the inertial solutions in everyday usage.

Recent emerging deep learning based inertial odometry technique, e.g. IONet \cite{Chen2018} is capable of extracting features and estimating continuous trajectories directly from raw inertial streams without any handcrafted engineering, outperforming previous model-based approaches in terms of accuracy and robustness. Other data-driven methods \cite{Yan2018} \cite{Cortes2018} learn to predict velocities in order to constrain system error drift, achieving competitive performance. There is growing interest in applying deep neural networks to learning motion from time-series data, due to its potential for model-free generalization.

These data-driven approaches require a significant amount data for training, validation and testing. In addition, highly precise labels (ground-truth values for locations, velocities, and orientations) are extremely important in supervised training. Most existing datasets aim at collecting sensor data from vehicles, in which installed IMUs are high accuracy devices (fiber gyro) and combined together with GPS, e.g. KITTI \cite{Geiger2013} and Oxford RobotCar \cite{Maddern2016}. Other datasets, e.g. Tum VI \cite{Schubert2018}, ADVIO \cite{Cortes2018-1} were collected for visual-inertial odometry with IMUs fixed in specific positions which does not reflect the challenges of everyday usage for example, a smartphone may be handheld, placed in a pocket or bag or even placed on a trolley. In these circumstances, the cameras are occluded. For conventional pedestrian dead reckoning, there is a lack of evaluation benchmarks, which impacts fair and objective comparison of different techniques. 

In this paper, we present and release the \textbf{Oxford Inertial Odometry Dataset (OxIOD)}, with abundant data (\textbf{158 sequences} and significant distances (\textbf{42.587 km} in total). For the majority of sequences, high precision labels with locations, velocities, and orientations are provided by an Optical Motion Capturing System (Vicon) \cite{Vicon2017} for training and evaluating models. Longer range sequences use a Google Tango visual-intertial odometry tracker \cite{Tango} to provide approximate ground-truth. These data were collected across different users/devices, various motions, and locations. We implement a deep learning algorithm and train and evaluate it to show the effectiveness of our proposed dataset. It is our hope that this dataset can boost objective research in learning based methods in inertial navigation.

\section{RELATED WORK}

A large amount of research work has been done in areas like autonomous driving and odometry, thus giving rise to many related datasets. Some datasets include visual and inertial data, and are used for developing visual-inertial odometry and SLAM algorithms. Others include datasets that focus on human gait and activity recognition. A brief overview of these datasets, as well as the comparison of these datasets to our dataset will be given in the following parts of this section. Inertial navigation using low-cost MEMS IMUs is briefly introduced.


\begin{table*}[ht]
\caption{Comparison of datasets with IMU and ground truth}
\label{dataset_compare}
\begin{center}
\begin{tabular}{p{1.8cm} p{0.5cm} p{1.4cm} p{1.5cm} p{1.2cm} p{2cm} p{2cm} p{2cm} p{1.2
cm}}
\hline
dataset & year & environment & carrier & IMUs & sample frequency & ground truth & ground truth accuracy & data size\\
\hline
KITTI & 2013 & outdoors & car & OXTS RT 3003 & 10 Hz & OXTS RT 3003& 10cm & 22 seqs, 39.2km\\
EuRoC MAV & 2016 & indoors & MAV & ADIS16488& 200Hz & laser tracker and motion capture & 1mm & 11 seqs, 0.9km\\
Oxford RobotCar & 2016 & outdoors & car & NovAtel SPAN-CPT& 50 Hz & fused GPS/IMU & unknown & 1010.46km\\
TUM VI & 2018 & in-/outdoors & human (handheld) & BMI160 & 200Hz & motion capture pose (only some parts) & 1mm & 28 seqs, 20km\\
ADVIO & 2018 & in-/outdoors & human (handheld) & iPhone 6s & 100Hz &  pseudo ground truth & unknown & 23 seqs, 4.5 km\\
\hline
\textbf{OxIOD (Ours)} & 2018 & indoors & human (handheld, pocket, trolley, bag) & InvenSense ICM-20600 & 100 Hz & Vicon & 0.5mm & \textbf{158 seqs}, \textbf{42.587km}\\
\hline
\end{tabular}
\end{center}
\end{table*}

\subsection{Inertial Odometry Datasets}

Table \ref{dataset_compare} shows representative datasets that include inertial data.
In KITTI \cite{Geiger2013}, Oxford RobotCar \cite{Maddern2016} and EuRoC MAV datasets \cite{Burri2016}, the sensors are rigidly fixed to the chassis, which is suitable for studying vehicle movements, but shows certain weakness in studying human movements. The TUM VI dataset \cite{Schubert2018} was collected to evaluate visual inertial odometry, with people holding the device in front of them. The ground truth in TUM VI is only provided at the beginning and ending of the sequences (in their vicon room), while during most of the trajectories there is no ground truth. Similarly, in ADVIO \cite{Cortes2018-1}, the dataset is rather short (4.5 km) and only offers pseudo ground truth generated by their handcrafted inertial odometry algorithm. In our dataset, the data collection device was placed in a hand, pocket, handbag and on the trolley separately and test subjects walk normally, slowly or run, as well as halting. Except for the large-scale localization subset (26 sequences, 3.465 km), the rest of our motion data (132 sequences, 39.122 km) were labelled with a very high-precise motion capture system with an accuracy of 0.5 mm in locations. In addition, inertial measurements from multiple users and devices were collected. Our dataset can better represent human motion in everyday conditions and thus has a greater diversity. 

As we can see from Table \ref{dataset_compare}, our dataset has a large amount of data from 158 sequences, leading to a total distance of 42.587km. The data size of OxIOD is larger than most other inertial odometry datasets, and hence is suitable for deep neural network methods, which require large amounts of data and high accuracy labels. It should be noted that the total length of the dataset even exceeds those collected by vehicles. 

\subsection{Gait and Activity Datasets}

There are also some datasets focusing on human gait and activities, which are somewhat similar to our dataset, but do not concentrate on tracking. Some of these datasets measure human activities, such as USC-HAD \cite{Zhang2012}, CMU-MMAC \cite{DeLaTorre2008}, and OPPORTUNITY \cite{Chavarriaga2013}. Though some of these datasets have inertial and accurate pose as ground truth, they can hardly be used to test odometry or SLAM algorithms, since the test subjects do not move much during the experiments.
Some other datasets, like MAREA \cite{Siddhartha2017}, focus on human gait recognition and collect inertial data while the carriers are walking or running. However, these datasets lack solid ground truth and thus limit their usage in testing odometry algorithms.

\subsection{Inertial Navigation Using Low-cost IMUs}

Due to high sensor noise and bias, and a lack of rigid body attachment, conventional Strapdown Inertial Navigation Systems (SINS) which integrate inertial measurements into orientations velocities and locations through state-space models are hard to implement on Low-cost MEMS IMUs. One of the common solutions is to combine IMU with a visual sensor to construct visual-inertial odometry (VIO) \cite{Li2013b} \cite{Leutenegger2015} \cite{Forster2017}, showing extremely good performance in tracking mobile robots, drones, and smart phones. However, VIOs require careful calibration/initialization and time synchronization of the inertial and visual sensors. VIOs will not work if its camera operates is encountered in low-light or featureless environments or even occluded. For example, in the context of indoor pedestrian navigation, the practical way to use a phone is to hold the device in a hand, place in a pocket or bag rather than pointing the camera in front of user's body. As an alternative solution, Pedestrian Dead Reckoning (PDR)s emerged to track pedestrian using inertial measurements only by detecting steps, estimating step length and heading to update locations \cite{Harle2013}. Recent research focuses fusing other sensor modalities with the PDR system to constrain system drifts, such as magnetic field distortions \cite{Wang2016a}, UWB \cite{LiuYDJLT17} or magneto-inductive fields~\cite{wei2018a}. The assumption of PDR systems is that the users exhibit periodic motion, which constrains their usage to situations where steps can be accurately detected. 

Recent emerging deep learning techniques prove that deep neural networks are capable of modeling high level motion directly from raw sequence data, such as DeepVO \cite{Wang2017} and VINet \cite{Clark2017a}. IONet \cite{Chen2018} proposes to formulate inertial odmetry as a sequential learning problem and constructs a deep recurrent neural network framework to reconstruct trajectories directly from raw inertial data, outperforming traditional model-based methods. Other learning-based learn velocities to constrain system drift \cite{Yan2018} \cite{Cortes2018}, or zero-velocity phase for inertial systems \cite{Wagstaff2018}. We implement the learning algorithms above for velocity/heading regression and for trajectory estimation. We also show that our dataset can be used to evaluate PDR models, demonstrating that it is a useful benchmark for any pedestrian tracking system, whether model-based or data-driven.

 \begin{figure}
    	\centering
        \includegraphics[width=0.5\textwidth]{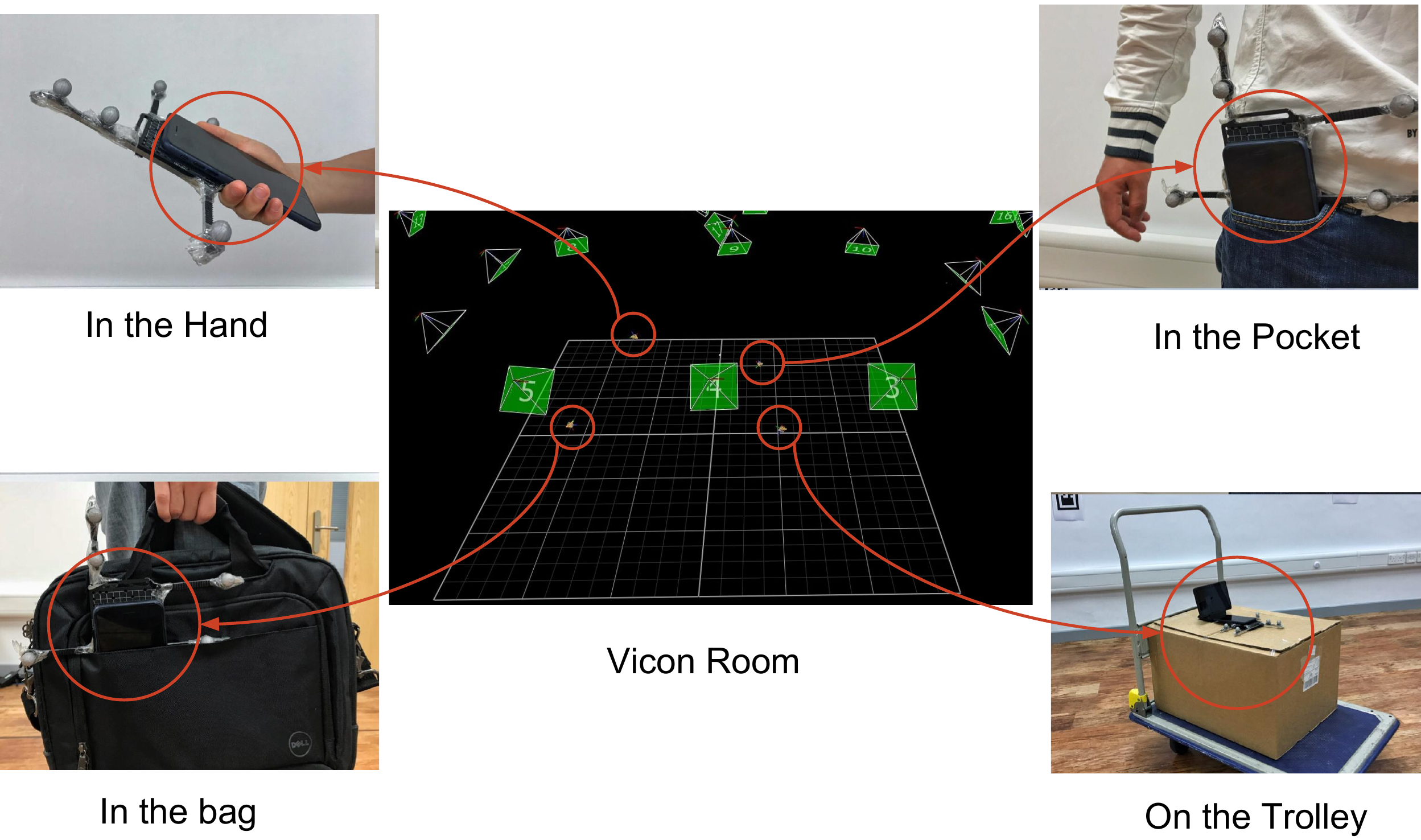}
        \caption{\label{fig:attachments} Inertial data are collected from a smartphone in four different attachments: handheld (left above), pocket (right above), handbag (left below), trolley (right below). With the help of Vicon System, high-precise motion labels can be provided.}
    \end{figure} 

\section{Sensor Setup}

The majority of the data in this dataset was collected with an iPhone 7 plus. We used IMUs to collect inertial (accelerometer and gyroscope) and magnetic field/compass data. All the sensors we used were integrated in the mobile phone. We used a Vicon motion capture system \cite{Vicon2017} to record ground truth.

		\begin{figure*}
    	\centering
        \begin{subfigure}[t]{0.19\textwidth}
        	\includegraphics[width=\textwidth]{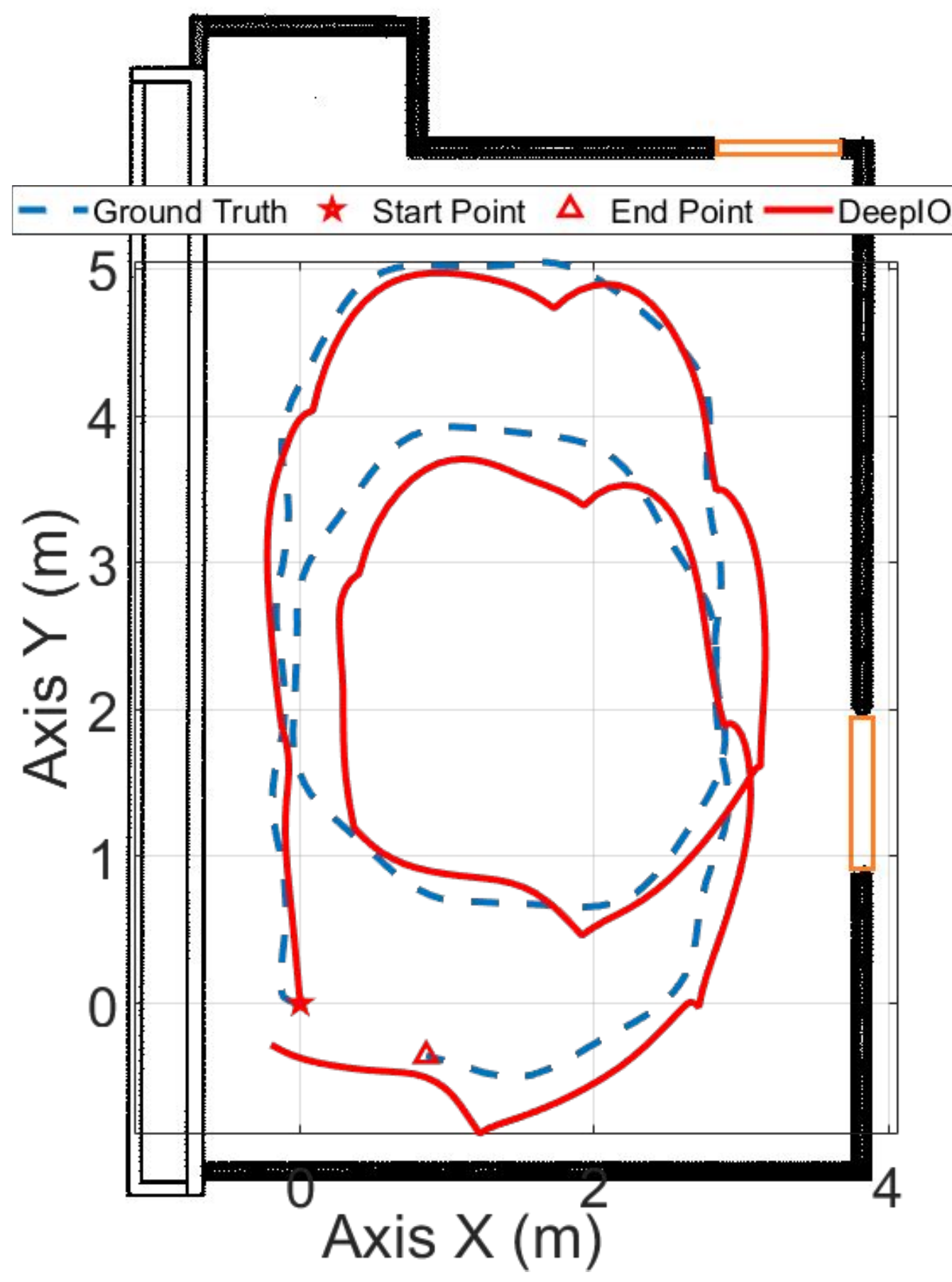}
        	\caption{\label{fig:vicon room} Vicon Room}
        \end{subfigure}
        \begin{subfigure}[t]{0.38\textwidth}
        	\includegraphics[width=\textwidth]{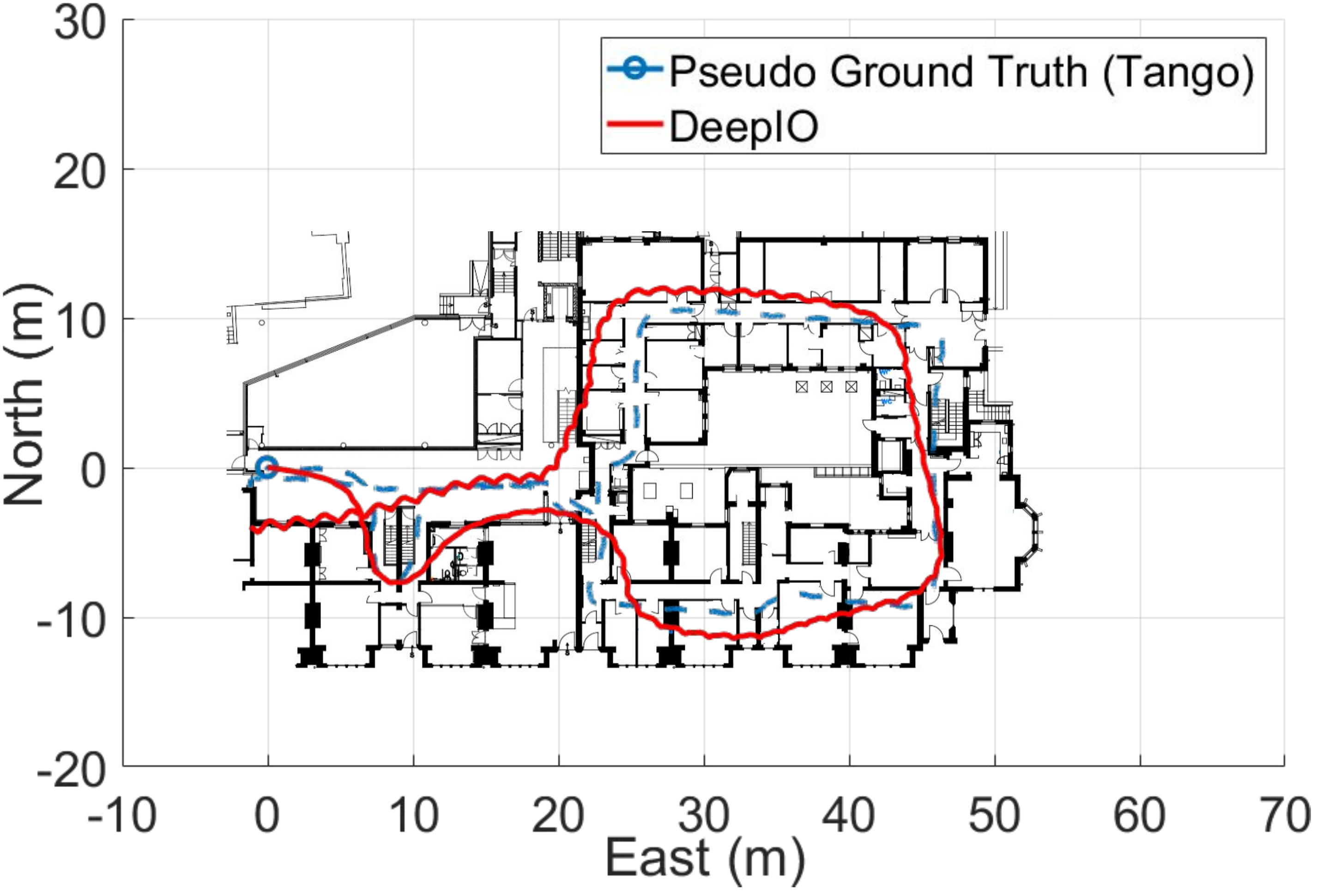}
        	\caption{\label{fig:floor1} Office Floor 1}
        \end{subfigure}
        \begin{subfigure}[t]{0.38\textwidth}
        	\includegraphics[width=\textwidth]{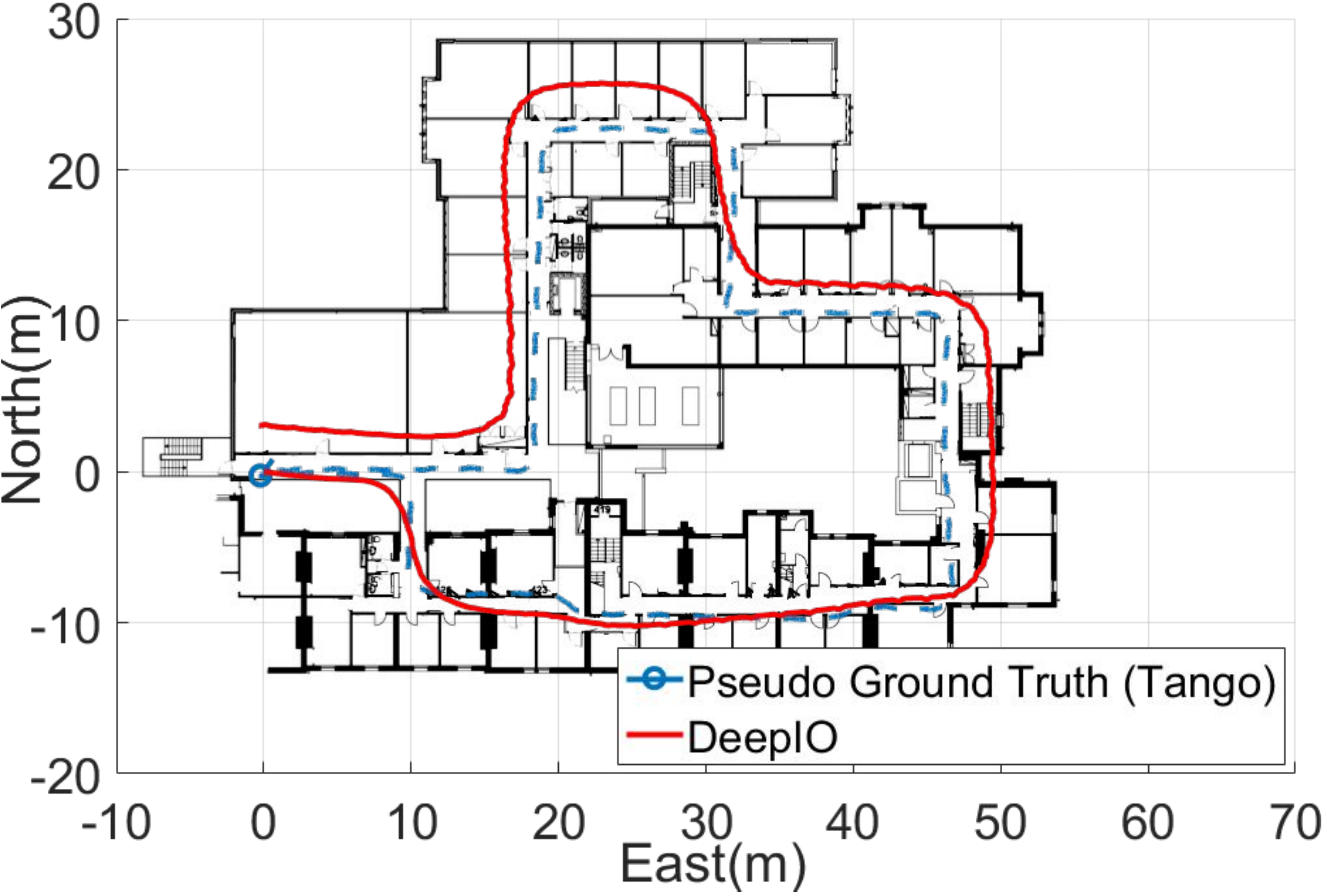}
        	\caption{\label{fig:floor2} Office Floor 2}
        \end{subfigure}
        \caption{\label{fig:map} The map illustration of a) the Vicon Room, b) Office Floor 1, and c) Office Floor 2. The ground truth for vicon room was captured by the Vicon System. The pseudo ground truth was provided by Google Tango for office floors. The DeepIO trajectories were generated by the learning model, which was implemented according to IONet \cite{Chen2018}, and trained above our proposed OxIOD dataset.}
    \end{figure*}

The other part of the dataset was collected with iPhone 6, iPhone 5 and Nexus 5. Large-scale localization was conducted on two office floors, where we used a Google Tango Tablet \cite{Tango} as pseudo ground truth, which will be introduced in latter sections. The sensor models of IMUs and magnetometers in these mobile phones are listed in Table \ref{sensor_list}.

\begin{table}[ht]
\caption{Sensors}
\label{sensor_list}
\begin{center}
\begin{tabular}{c c c}
\hline
Mobile Phone & IMU & Magnetometer\\
\hline
iPhone 7 Plus & InvenSense ICM-20600 & Alps HSCDTD004A\\
iPhone 6 & InvenSense MP67B & AKM 8963\\
iPhone 5 & STL3G4200DH &  AKM 8963\\
Nexus 5 & - & - \\
\hline
\end{tabular}
\end{center}
\end{table}

\subsection{IMU}

The IMU inside iPhone 7 Plus is an InvenSense ICM-20600, which is a 6-axis motion tracking device. It combines a 3-axis gyroscope and a 3-axis accelerometer. 16-bit ADCs are integrated in both gyroscope and accelerometer. The sensitivity error of the gyroscope is $1\%$, and the noise is $4mdps/\sqrt{Hz}$. The accelerometer noise is 100$\mu g/\sqrt{Hz}$.

\subsection{Magnetometer}

The Alps HSCDTD004A in iPhone 7Plus is a 3-axis geomagnetic sensor, which is mainly used for electronic compasses. It has a measurement range of $\pm 1.2 mT$ and an output resolution of $0.3\mu T/LSB$.

\subsection{Vicon System}

We used 10 Bonita B10 cameras in the Vicon Motion Tracker system \cite{Vicon2017}, which circles the area where we did the experiments. The Bonita B10 camera has a frame rate of 250 fps, and resolution of 1 megapixel (1024*1024). The lens operating range of Bonita B10 can be up to 13 m. These features enable us to capture ground truth data with a precision down to 0.5 mm, making the ground truth very accurate and reliable. The software we used in the Vicon system is Vicon Tracker 2.2. We connect the Vicon motion tracker to ROS with vicon\_bridge, and recorded the data stream with rostopic.

\subsection{Time Synchronization}

The IMU and magnetometer are integrated in the mobile phone, so they share the same time stamp. 
Vicon data recorded with ROS is also saved with timestamp. Before each experiment, we synchronize the time of iPhone 7 Plus and ROS with UTC, and thus all the timestamp recorded with the data will be synchronized.

\section{OxIOD Dataset}
In this section, we introduce OxIOD - the proposed inertial tracking dataset, with various attachments, motion modes, devices and users to present the sensory readings in everyday usage, illustrated in Table \ref{tb:dataset}. Our dataset has \textbf{158 sequences}, and the total walking distance and recording time are \textbf{42.5 km}, and \textbf{14.72 h} (53022 seconds). We collected the sensor data with off-the-shelf consumer phones, and provided very high-precise ground truth with the aid of Motion Caption system (Vicon). For large-scale localization data, we instead used the Tango device (visual-inertial odometry) to offer pseudo ground truth.

	\begin{figure*}
    	\centering
        \begin{subfigure}[t]{0.48\textwidth}
        	\includegraphics[width=\textwidth]{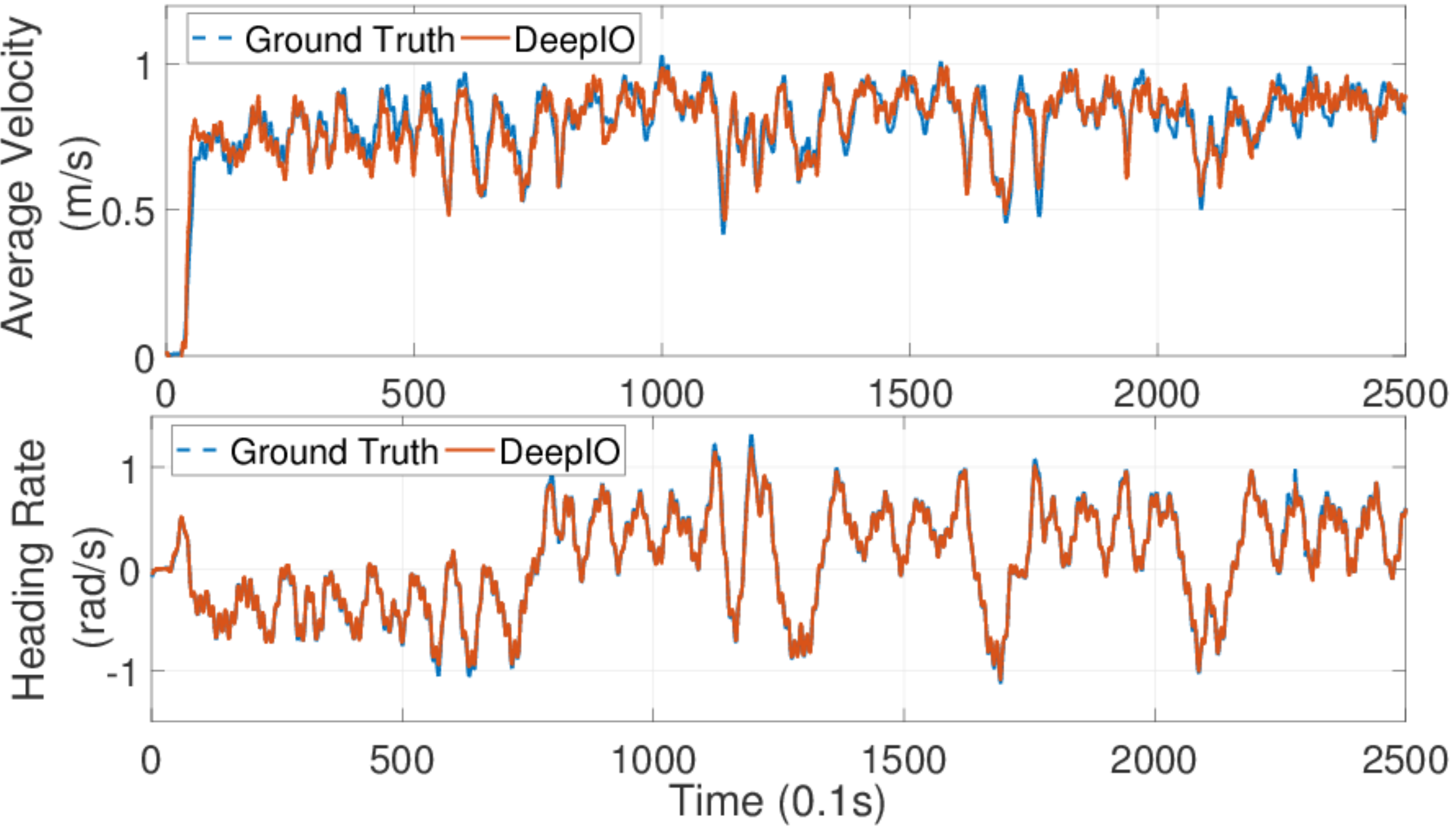}
        	\caption{\label{fig:speed_normal} Walking Normally}
        \end{subfigure}
        \begin{subfigure}[t]{0.48\textwidth}
        	\includegraphics[width=\textwidth]{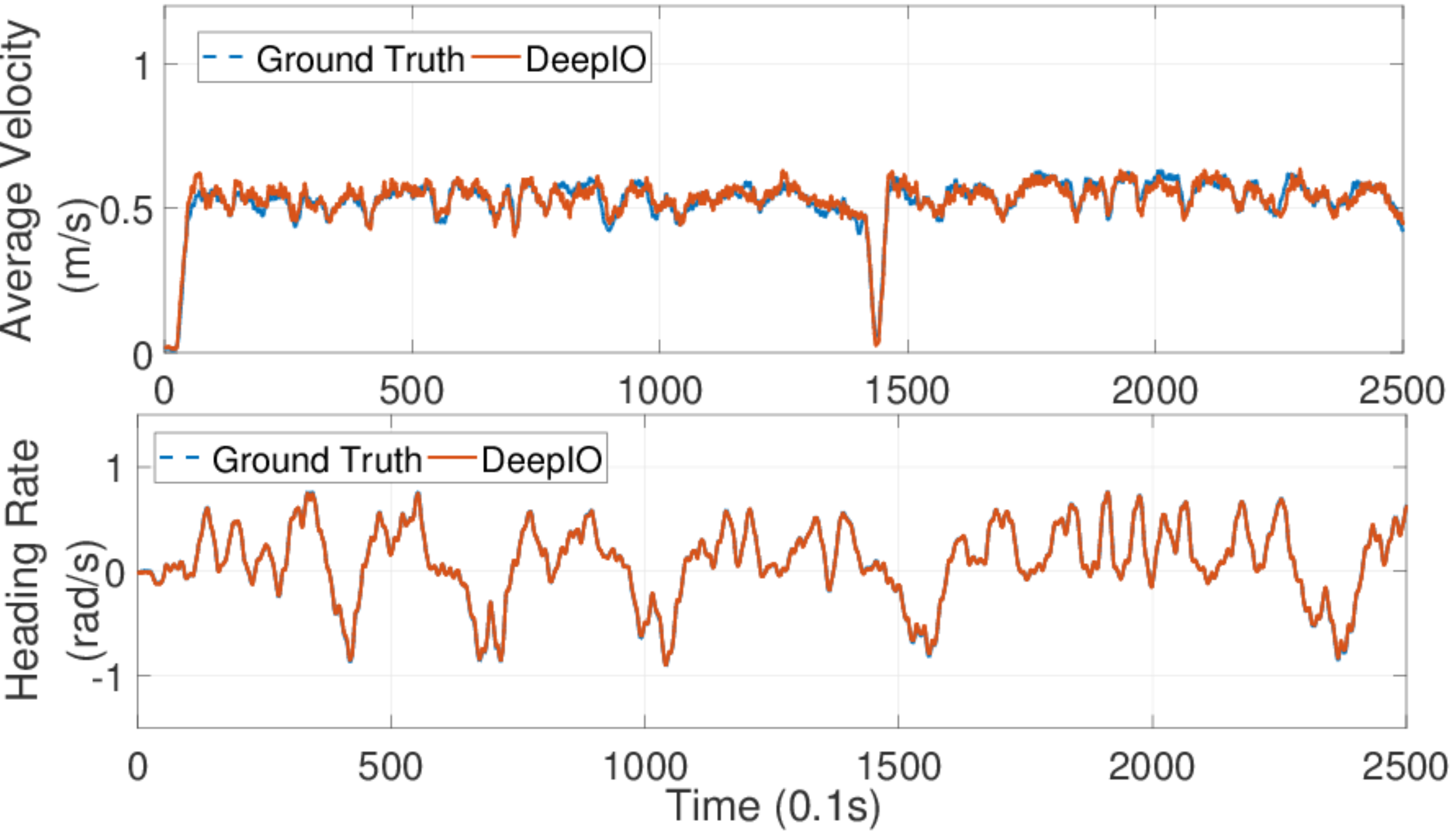}
        	\caption{\label{fig:speed_slow} Walking Slowly}
        \end{subfigure}
        \begin{subfigure}[t]{0.48\textwidth}
        	\includegraphics[width=\textwidth]{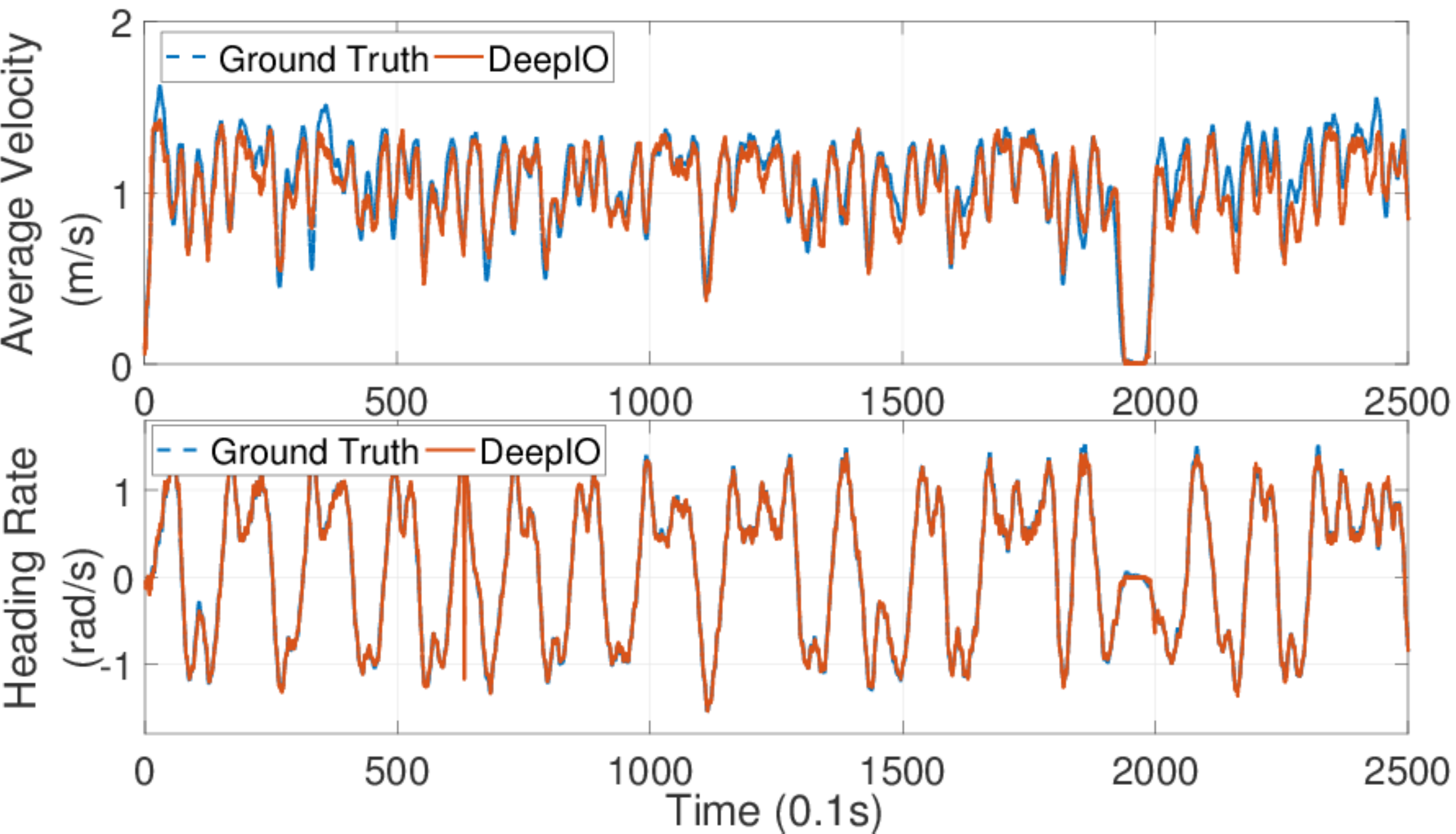}
        	\caption{\label{fig:speed_run} Running}
        \end{subfigure}
        \begin{subfigure}[t]{0.48\textwidth}
        	\includegraphics[width=\textwidth]{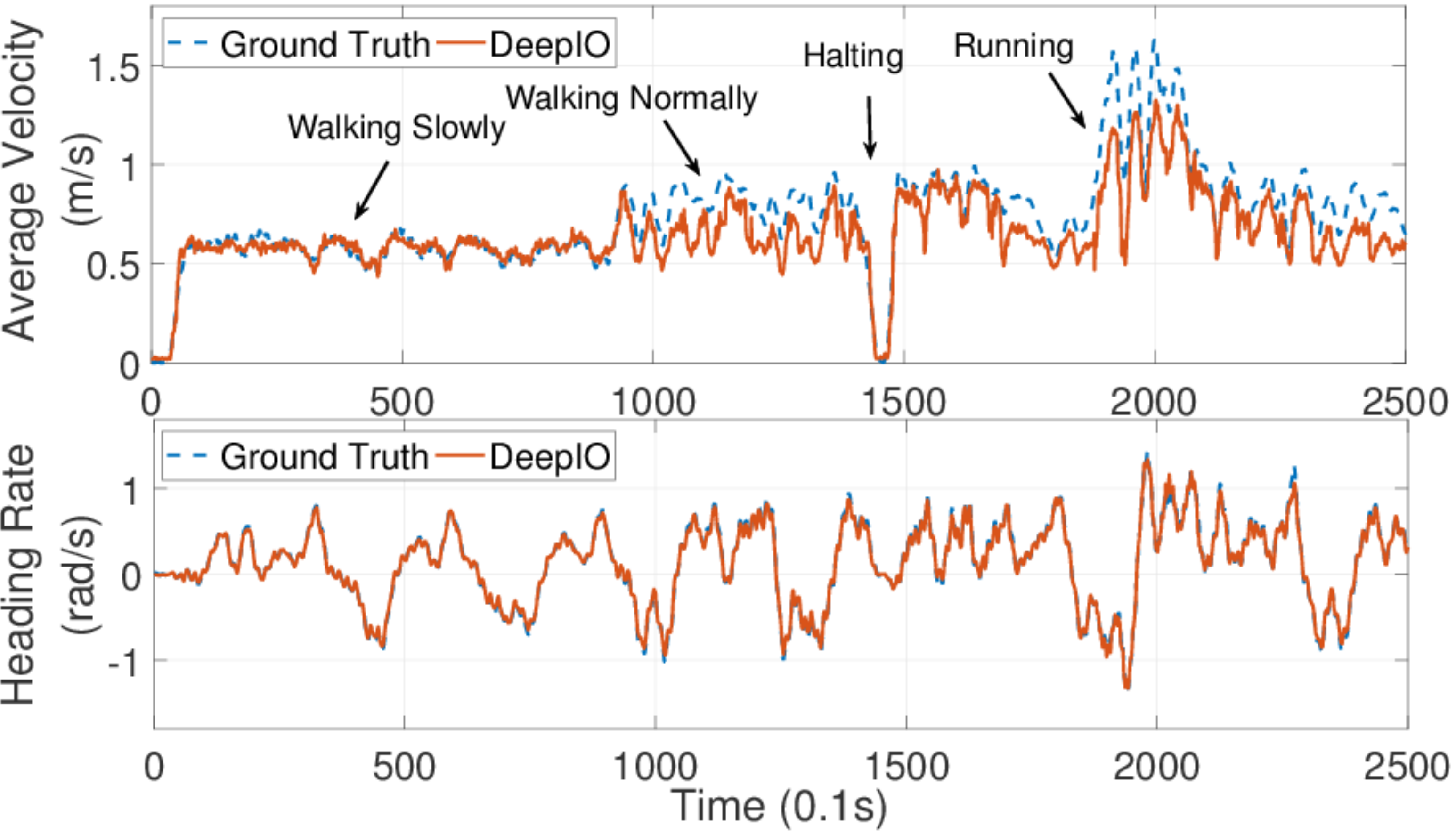}
        	\caption{\label{fig:speed_mix} Mixed Activities}
        \end{subfigure}
        \caption{\label{fig:velocity estimation} The velocity and heading estimation for a) walking normally, b) walking slowly, c) running and d) mixed motion modes. The ground truth was captured by Vicon System, while the values from DeepIO were predicted by the learning model trained on our proposed dataset.}
    \end{figure*}

\begin{table}[ht]
\caption{OxIOD Dataset}
\label{tb:dataset}
\begin{center}
\begin{tabular}{c c c c c}
\hline
~ & Type & Seqs & Time (s) & Distance (km) \\
\hline
\multirow{4}*{Attachments} & Handheld & 24 & 8821 & 7.193\\
~ & Pocket & 11 & 5622 & 4.231\\
{\tiny (iPhone 7P/User 1)} & Handbag & 8 & 4100 & 3.431\\
{\tiny (Normally Walking)} & Trolley & 13 & 4262 & 2.685\\
\hline
\multirow{3}*{Motions} & Slowly Walking & 8 & 4150 & 2.421\\
~ & Normally Walking & -  & - & - \\
~ & Running & 7 & 3732 & 4.356\\
\hline
\multirow{3}*{Devices} & iPhone 7P & - & - & - \\ 
~ & iPhone 6 & 9 & 1592 & 1.381\\
~ & iPhone 5 & 9 & 1531 & 1.217\\
~ & Nexus 5 & 8 & 4021 & 2.752\\
\hline
\multirow{4}*{Users} & User 1 & - & - & - \\ 
~ & User 2 & 9 & 2928 & 2.422\\
~ & User 3 & 7 & 2100 & 1.743\\
~ & User 4 & 9 & 3118 & 2.812\\
~ & User 5 & 10 & 2884 & 2.488\\
\hline
\multirow{2}*{Large Scale} & floor 1 & 10 & 1579 & 1.412\\
~ & floor 2 & 16 & 2582 & 2.053\\
\hline
Total &  & 158 & 53022 & 42.587\\
\hline
\end{tabular}
\end{center}
\end{table}

\subsection{Attachments}

Most of the existing tracking datasets assume that the sensors are fixed. The sensors are installed at specific position in autonomous cars or UAV, for example, in the KITTI \cite{Geiger2013}, Oxford Robotcar \cite{Maddern2016},  EuRoC MAV \cite{Burri2016} dataset, or the devices are held by hand with the camera pointing toward the walking direction, for example, in TUM VI \cite{Schubert2018}, ADVIO \cite{Cortes2018-1}. But in real application scenarios, the IMUs experience the distinct motions when attached in different places, and there is uncertainty related with how users may use their phone. We aim to reflect the every day usage by selecting four common situations, e.g. handheld, in the pocket, in the handbag and on the trolley. The data was collected by iPhone 7Plus with a pedestrian (user 1), carrying phone naturally and walking normally inside a room installed with motion capture system. Figure \ref{fig:vicon room} shows the map illustration of Vicon Room (5 m * 8 m).

	\begin{figure*}
    	\centering
        \begin{subfigure}[t]{0.24\textwidth}
        	\includegraphics[width=\textwidth]{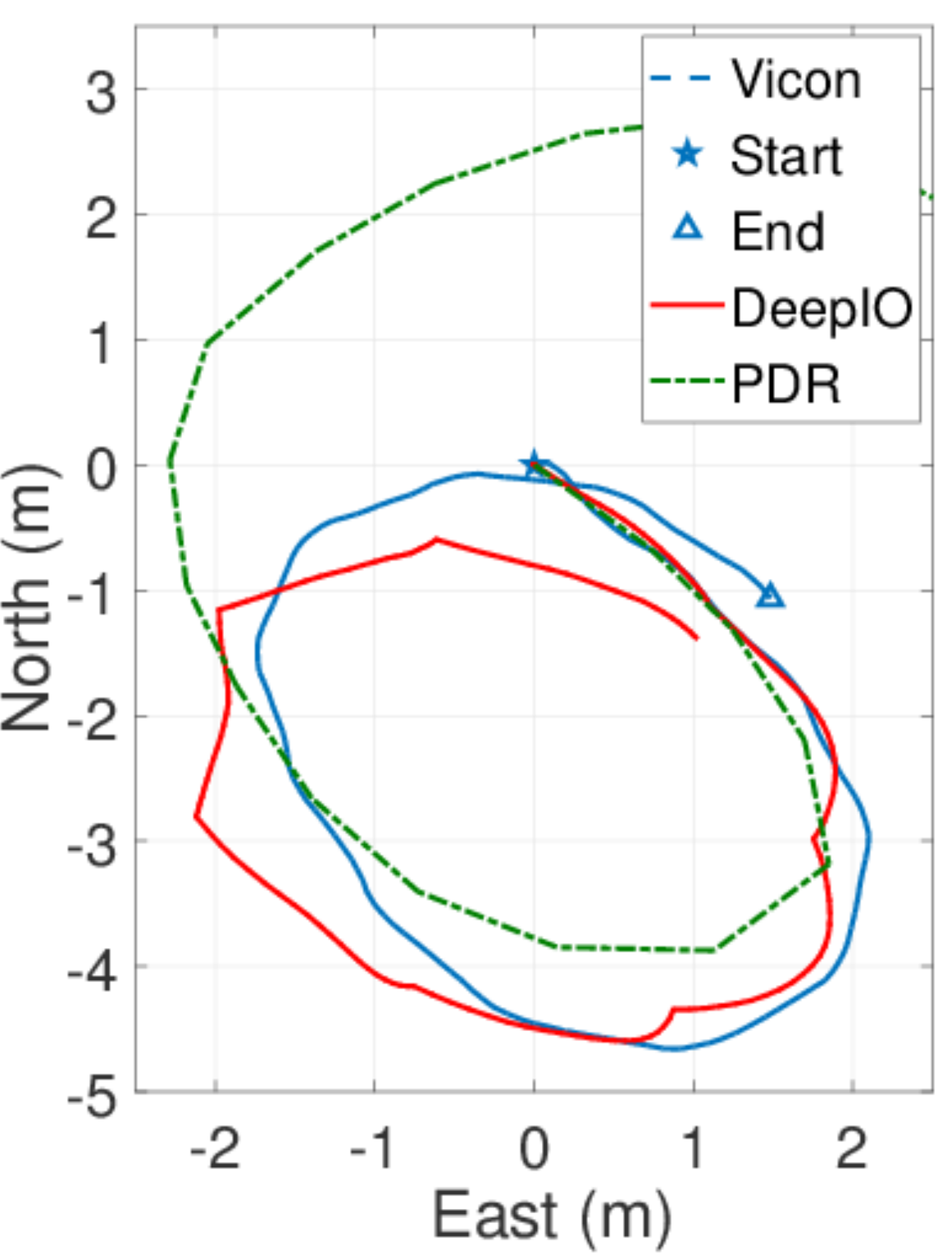}
        	\caption{\label{fig:handheld} Handheld}
        \end{subfigure}
        \begin{subfigure}[t]{0.24\textwidth}
        	\includegraphics[width=\textwidth]{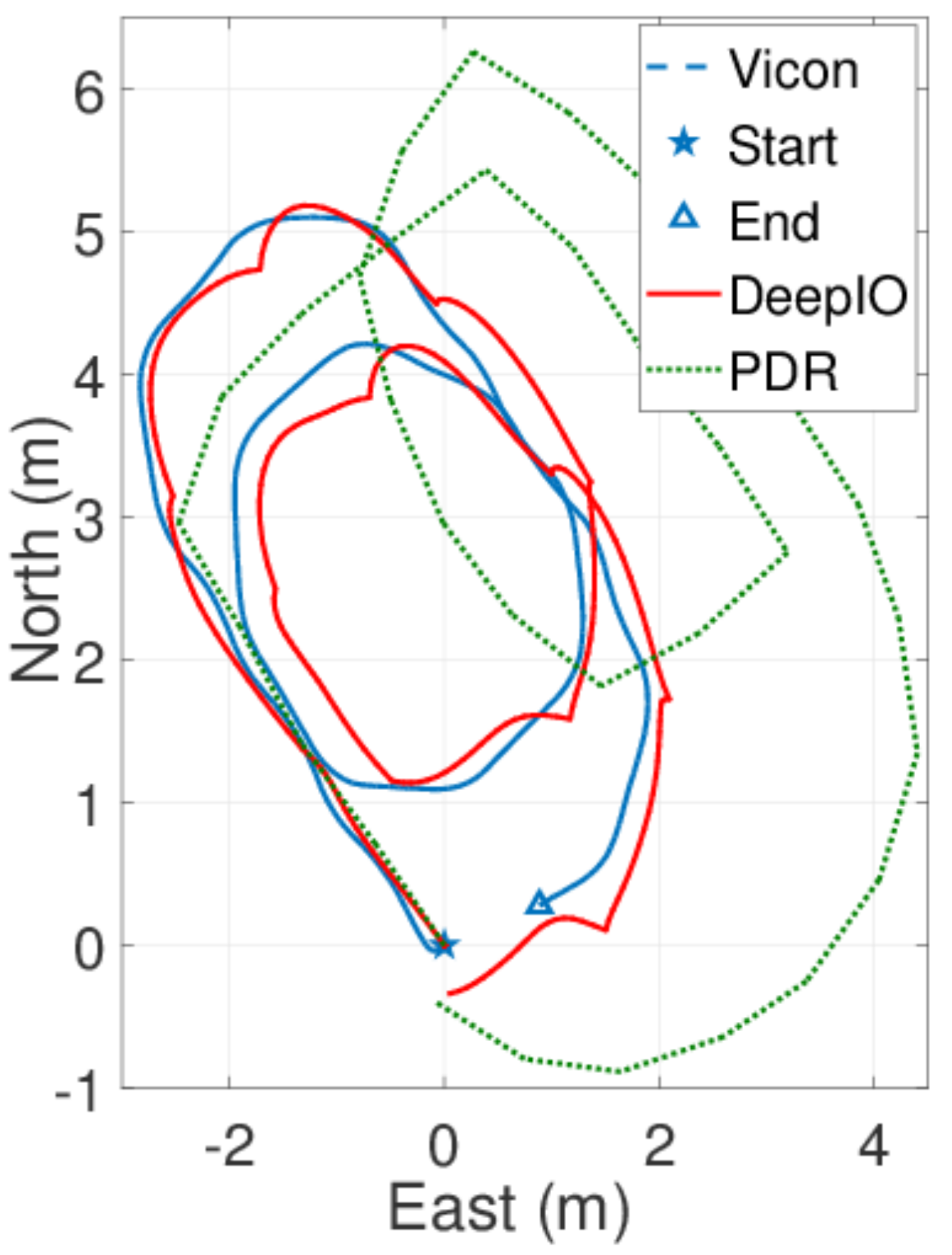}
        	\caption{\label{fig:pocket} In Pocket}
        \end{subfigure}
        \begin{subfigure}[t]{0.24\textwidth}
        	\includegraphics[width=\textwidth]{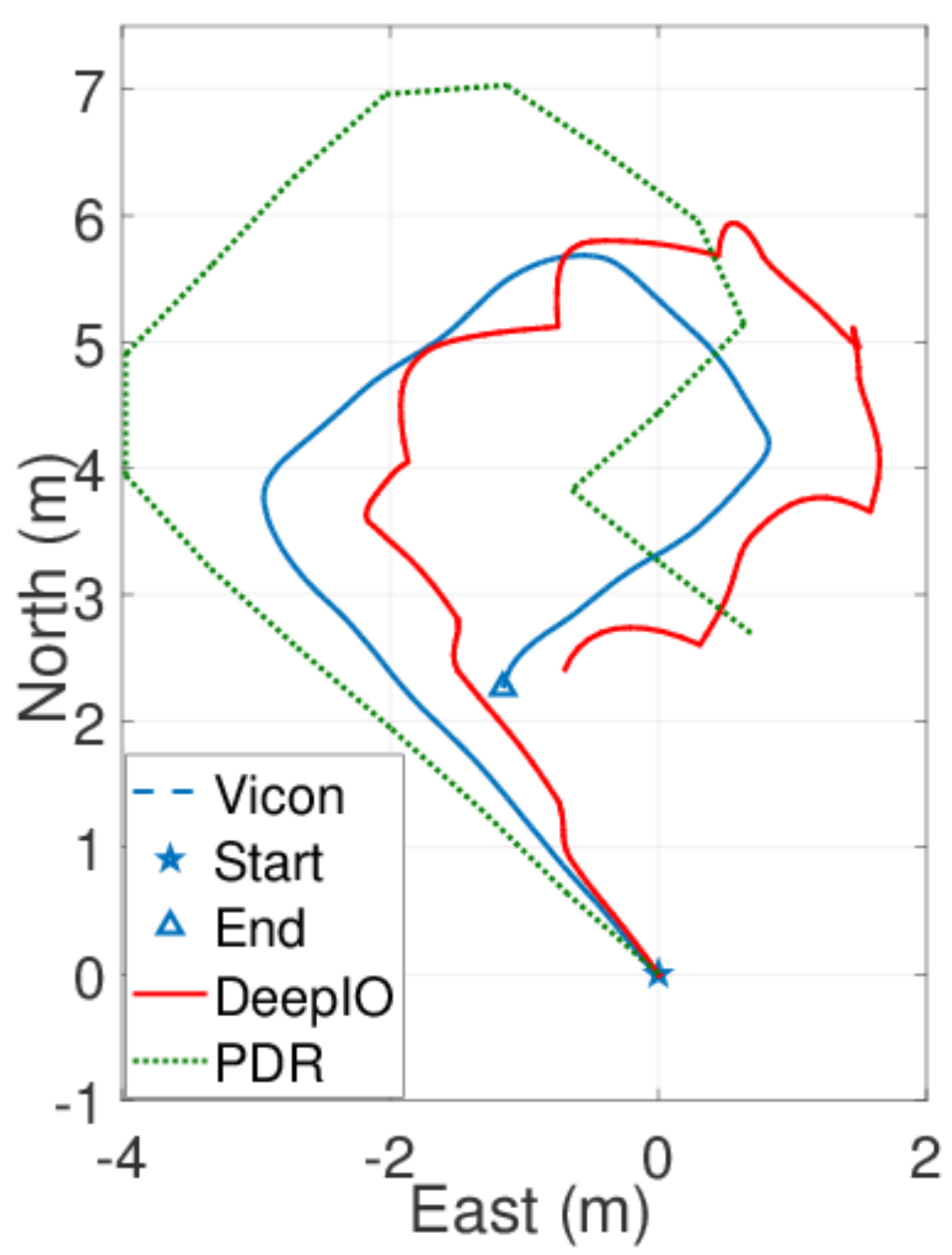}
        	\caption{\label{fig:handbag} In Handbag}
        \end{subfigure}
        \begin{subfigure}[t]{0.24\textwidth}
        	\includegraphics[width=\textwidth]{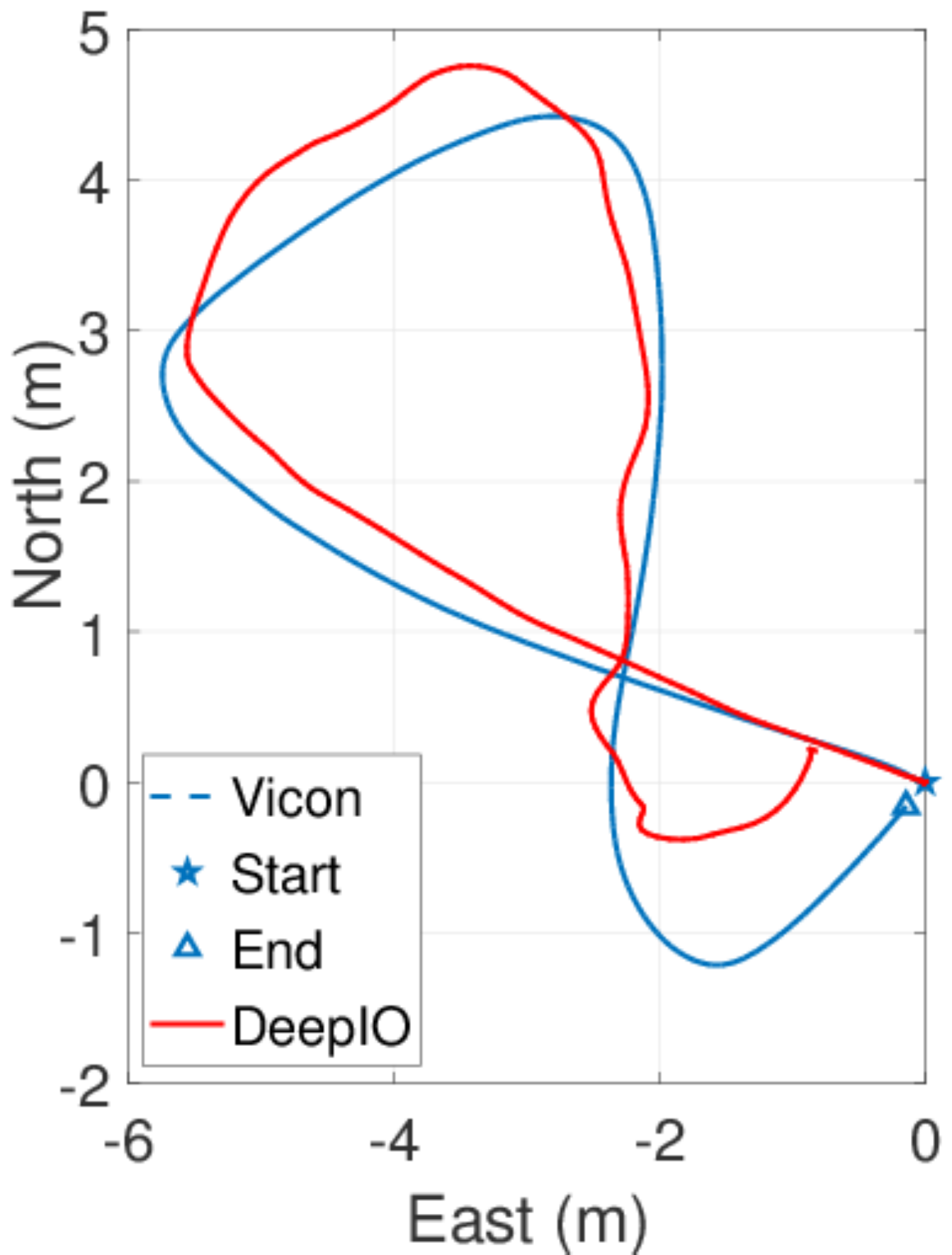}
        	\caption{\label{fig:trolley} On Trolley}
        \end{subfigure}
        \caption{\label{fig:deep tracking} The trajectories reconstruction for pedestrian tracking with device in four attachments: a) in the hand, b) in the pocket, c)in the handbag, and d) on the trolley respectively. We implemented the DeepIO (Deep Learning based Inertial Odometry) according to IONet \cite{Chen2018}, and a simple PDR algorithm for comparison. PDRs do not work when the device was placed on the trolley, as no step can be detected in this situation. The ground truth values are provided by the Vicon System.}
    \end{figure*}

\subsection{Motion Modes}
Humans move in different motion modes in their everyday activities. We selected and collected data from four typical motion models: halting, walking slowly, walking normally and running. All experiments were performed by User 1 with iPhone 7Plus in hand.

\subsection{Devices and Users}
Both the sensors used and the walking habits of users impact the performance of inertial tracking systems. In order to ensure inertial tracking invariant across devices and users, we collected data from several devices and users. Four off-the-shelf smartphone were selected as experimental devices: iPhone 7Plus, iPhone 6, iPhone 5, and Nexus 5, as shown in Figure \ref{sensor_list}. Five participants performed experiments to walk with phone in hand, pocket and handbag respectively. The mixed data from various devices and users can also be applied in classifying devices and users.

\subsection{Large-scale localization}
Besides the extensive data collection inside the VICON room, we also conduct large-scale tracking in two environments. Without the help of Vicon system, the Google Tango device was chosen to provide pseudo ground truth. The participant was instructed to walk freely in a office building on two separate floors (about 1650 $m^2$ and 2475 $m^2$). The smartphones were placed in the hand, pocket and handbag respectively, but the tango device was attached on body to capture the precise trajectory on the chest. The floor maps with pseudo ground truth trajectories captured by Google Tango and generated trajectories from learning models are illustrated in Fig. \ref{fig:floor1} and Fig. \ref{fig:floor2}. 


\section{Experiments}

In this section, we trained deep neural network models on our proposed OxIOD dataset, and conducted experiments to evaluate them. The results from deep learning based inertial odometry are presented as DeepIO in short.

\subsection{Velocity and Heading Regression}
\label{sec: velocity}

As a demonstration of training performance, a recurrent neural network (RNN) model was trained to predict the average velocity and heading rate of pedestrian motion. The average velocity $\bar{v}$ and heading rate $\dot{\psi}$ are defined as the location displacement $\Delta l$ and heading change $\Delta \psi$ during a window size of time $n$:
	\begin{equation}
		(\bar{v}, \dot{\psi}) = (\Delta l / n, \Delta \psi / n) .
	\end{equation}
In our experiment setup, the window size $n$ was selected as 2 seconds, so a sequence of inertial data (200 frames) $(\{(\mathbf{a}_i, \mathbf{w}_i)\}_{i=1}^{n})$ is fed into RNN to predict the average velocity $\bar{v}$ and heading rate $\dot{\psi}$:
	\begin{equation}
		(\bar{v}, \dot{\psi}) = \text{RNN} (\{(\mathbf{a}_i, \mathbf{w}_i)\}_{i=1}^{n}) .
	\end{equation}
Considering the long time dependencies of the input time-series data (200 frames), we choose one-layer Long Short Term Memory (LSTM) as the framework with the hidden states ($h = 128$) \cite{Greff2016}. To prevent overfitting, a dropout layer of 0.25 is used after the LSTM layer. Finally, a fully-connected layer is defined to map the last output of LSTM into the average velocity and heading rate of the sequence. The framework is implemented through PyTorch.

The training data are from the walking normally (handheld, 24 seqs), walking slowly (8 seqs) and running (7 seqs) in the motion modes dataset. By minimizing the mean square error between the estimated values and ground truth provided by our dataset, the optimal parameters inside the RNN are recovered via the ADAM optimizer \cite{Kingma2014} with a learning rate of 0.0001. To test its generalization ability, we perform randomly walking in the Vicon Room, and use the trained neural network to predict the values for selected three motion modes respectively and a mix of activities. From Fig. \ref{fig:velocity estimation}, it can be seen that the trained RNN can model the motions well, and generalize well to mixed activities. 

\subsection{Training and Evaluating Deep Inertial Odometry}

We implement a learning based method (IONet framework \cite{Chen2018}: DeepIO) to reconstruct 2D trajectories from the raw inertial data. The polar displacement vector $(\Delta l, \Delta \psi)$ was predicted by a two-layer Bi-directional LSTM framework with hidden states $h=128$, using a sequence of inertial data $(\{(\mathbf{a}_i, \mathbf{w}_i)\}_{i=1}^{n})$ during a window size of time $n$:
	\begin{equation}
		(\Delta l, \Delta \psi) = \text{RNN}(\{(\mathbf{a}_i, \mathbf{w}_i)\}_{i=1}^{n}) .
	\end{equation}
Subsequently the current location $(L^x, L^y)$ is updated by
	\begin{equation}
     	\label{eq: location}
    	\left\{
    	\begin{aligned}
    		L^x_n=L^x_0+\Delta l \cos(\psi_0+\Delta \psi) \\
        	L^y_n=L^y_0+\Delta l \sin(\psi_0+\Delta \psi),
        \end{aligned}
       \right.
    \end{equation}
where $(L^x_0, L^y_0)$, and $\psi_0$ are the initial location and heading of the sequence. 
We selected the handheld (24 seqs), pocket (11 seqs), handbag (8 seqs) and trolley (13 seqs) data from the attachments dataset to train the model. Other training details follow the same as in Section \ref{sec: velocity}.

We also perform random walking inside the Vicon Room with smartphone in four attachments, and these walking trajectories never show up in the training dataset. Fig. \ref{fig:deep tracking} illustrates the generated trajectories from the learning model. A PDR algorithm implemented according to \cite{Xiao2014} is compared with the DeepIO and the ground truth.

\section{Conclusions and Discussions}

We present and release an inertial odometry dataset (OxIOD) for training and evaluating learning-based inertial odometry models. Our dataset is much larger than previous inertial datasets with 158 sequences and 42.587 km in total distance, provided with high-precision labels recording locations, velocities and orientations. OxIOD has greater diversity than existing ones, with the data collected from different attachments, motion modes, users, devices and locations.

With release of this large-scale diverse dataset, it is our hope to boost research in the field of deep inertial navigation and localization, and enable future research in enhancing intelligence and mobility for mobile agents through long-term ubiquitous ego-motion estimation. 
Future work would include collecting data from more challenging situations, for example, 3D tracking. We also plan to create on-line common benchmark and tools for the comparison of odometry models. 









\bibliographystyle{IEEEtran}
\bibliography{refs.bib}

\end{document}